\documentclass[letterpaper]{article} 

\usepackage{cvpr} 
\usepackage{times}
\usepackage{epsfig}
\usepackage{graphicx}
\usepackage{amsmath}
\usepackage{amssymb}
\usepackage{algorithm}
\usepackage{caption,subcaption}
\usepackage{amsmath}
\usepackage[square,sort,comma,numbers]{natbib}
\usepackage{algpseudocode}

\DeclareMathOperator*{\argmax}{arg\,max}

\algtext*{EndWhile}
\algtext*{EndFor}
\algtext*{EndIf}

\makeatletter
\def\BState{\State\hskip-\ALG@thistlm}
\makeatother
\newcommand\numberthis{\addtocounter{equation}{1}\tag{\theequation}}

\usepackage[breaklinks=true,bookmarks=false]{hyperref}

\newcommand*\samethanks[1][\value{footnote}]{\footnotemark[#1]}

\cvprfinalcopy 

\def\cvprPaperID{****} 

\begin{document}

\title{Higher Order Mutual Information Approximation for Feature Selection}

\author{Jialin Wu\thanks{Authors contribute equally}\\
	Tsinghua University\\
	Dept. of Automation Engineering\\
	{\tt\small wujl13@mails.tsinghua.edu.cn}
	\and
	Soumyajit Gupta\samethanks\\
	University of Texas Austin\\
	Dept. of Computer Science\\
	{\tt\small smjtgupta@utexas.edu}
	\and
	Chandrajit Bajaj\\
	University of Texas Austin\\
	Dept. of Computer Science\\
	{\tt\small bajaj@cs.utexas.edu}
}

\maketitle

\begin{abstract}
   Feature selection is a process of choosing a subset of relevant features so that the quality of prediction models can be improved. An extensive body of work exists on information-theoretic feature selection, based on maximizing Mutual Information (MI) between subsets of features and class labels. The prior methods use a lower order approximation, by treating the joint entropy as a summation of several single variable entropies. This leads to locally optimal selections and misses multi-way feature combinations. We present a higher order MI based approximation technique called Higher Order Feature Selection (HOFS). Instead of producing a single list of features, our method produces a ranked collection of feature subsets that maximizes MI, giving better comprehension (feature ranking) as to which features work best together when selected, due to their underlying interdependent structure. Our experiments demonstrate that the proposed method performs better than existing feature selection approaches while keeping similar running times and computational complexity.
\end{abstract}

\section{Introduction}

\footnotetext{This work is supported by the NIH Grants \#R41 GM116300, \#R01 GM117594.}
\footnotetext{Primary Contact: C. Bajaj, ; email: bajaj@cs.utexas.edu; 201 East 24th Street, POB 2.324A, 1 University Station, C0200, Austin, TX, 78712-0027.}

Feature selection is a dimensionality reduction technique for a wide range of search problems. A common practice used in machine learning is to find subset of available features for a learning algorithm. The best subset contains the least number of dimensions that most contribute to accuracy~\cite{ladha2011feature}. Feature selection differs from standard dimensionality reduction. Both methods seek to reduce the number of attributes in the dataset, but a dimensionality reduction method does so by creating new combinations of attributes, where as feature selection methods includes and excludes attributes in the data without changing them.

The central premise when using a feature selection technique is that the data contains many features that are either \textit{redundant} or \textit{irrelevant}, and can thus be removed without incurring much loss of information~\cite{bermingham2015application}. Redundant features refer to those having information which is already contained in other features, while irrelevant features are those with no useful information related to the class variable. Feature selection mostly acts as a filter, muting out features that are not useful in addition to the existing features , and thus avoid over-fitting, speed up computation, or improve the interpretability of the results.

Feature selection techniques can be broadly classified into two groups: classifier dependent (\textit{wrapper} \cite{kohavi1997wrappers}, \textit{embedded} \cite{lal2006embedded}) and classifier independent \textit{filter}~\cite{kira1992practical}. Classifier dependent methods exploit the label information as a measure to guide the feature selection process. Examples are Backward Elimination and Forward selection. In forward selection, variables are progressively incorporated into larger subsets, whereas in backward elimination one starts with the set of all variables and progressively eliminates the least promising ones~\cite{guyon2003introduction}. The inherent advantage of this class of methods are that multi-collinearity issues are automatically handled. However, no prior knowledge about the actual relationship between the variables can be inferred. 

Classifier independent methods, on the other hand, define a scoring function between features and labels in the selection process. They are based only on general features like the correlation with the variable to predict. Filter methods suppress the least interesting variables and are particularly effective in computation time and robust to over-fitting. A very simple \textit{filter} type feature selection algorithm is a basic correlation analysis~\cite{hall2000correlation}, where only attributes which are correlated to the label are be chosen in a predictive analytics model. Correlation analysis helps to identify these ``features" right at the outset and allows a better understanding of how the attribute affects the predicted variable. However correlation table needs to be generated and collinear variables must be eliminate ``by hand".

The Mutual Information MI (also called cross entropy or information gain) is a widely used information theoretic measurement for the stochastic dependency of discrete random variables. MI based feature selection works similar to a standard correlation analysis, except that it is more robust when the variables are noisy or non-linearly related to the predicted or target variable. Heuristic greedy algorithms are adopted to find out the most relevant and least redundant feature at each step. Brown \etal~\cite{brown2012conditional} show that the redundant term should refer to conditional mutual information and summarizes a uniform framework for information theoretic feature selection. However, due to the computational cost, the joint probability density functions (pdf) are often approximated as multiplication of single variable distribution functions. This is based on two inconsistent independence assumptions in Gao \etal~\cite{gao2016variational}, which can be seen as first order approximation of Information Gain, leading to the miss of multi-way feature combinations. This feature independence assumption produces singleton feature elements, thereby failing to capture he joint MI gain over a collection of correlated variables. A higher order approximation is thus needed to counter this scenario and get a better comprehension of the inter-feature dependence.

The ICA (Independent Component Analysis) algorithm, based on infomax or maximum likelihood estimation, tries to represent an input vector $x = ( x_1,x_2,...,x_M)^T$ as a linear combination of independent signal vector $s = ( s_1,s_2,...,s_M)^T$ e.g. $X = AS$, where each $x_i$ and $s_i$ represents a random variable and $A$ is called the mixing matrix~\cite{comon1994independent}. The connection between ICA algorithm and information theory has been well known for many years~\cite{li2015information}. However, to our knowledge, most of work concerns extracting signal vector $s$ by finding out the optimal unmixing matrix ($W \sim A^{-1}$), that can minimize the mutual information(MI) between $s_i$, which leads to large running time.

To address all these issues, this paper presents a feature selection method, termed Higher Order Feature Selection (HOFS). This integrates ICA to approximate the MI in higher order dimension. Instead of producing a single list of features, we provide several subsets of features that maximizes MI, giving better comprehension as to which features work best when selected together due to their underlying interdependent  structure. This addresses Feature Ranking, by scoring the feature subsets based on their combined predictive power, which is a significant advantage over lower order models. Our model also considers simplifying the computational cost of calculating multi-variable MI using ICA. We compare the results of our experiments against well known feature selection algorithms on publicly available datasets. Observations show that we perform at par if not better than them in classification error and global MI captured using the subset of selected features, while keeping similar running times and computational complexity.

Rest of the paper is organized as follows: Sec.~\ref{sec2} sets up the basic framework of MI based feature selection approach with notations, definitions and prior work. Sec.~\ref{sec3} discusses the lo-order and ICA issues which we are trying to address in this paper. In Sec.~\ref{sec4}, we introduce the proposed Higher Order Feature Selection (HOFS) approach with its computation analysis and running times. All experiments on publicly available real datasets and their comparison results are illustrated in Sec.~\ref{sec5}. Experiments performed on synthetic data (numerical graphical models and heterogeneous feature models) for model checking are tabulated in Sec.~\ref{sec6}. We conclude our findings and discuss some of the future problems that can be addressed in Sec.~\ref{sec7}. 

\section{Feature Selection Background} \label{sec2}

\subsection{Notation}

Let $x$ denote a possible value of the discrete random variable (r.v.) $X$ which is drawn from the alphabet set $\mathcal{X}$. Also, let $p(X)$ define the probability density function (pdf) of $X$, and $P(X=x)$ denote the probability that r.v. $X$ takes value $x$. When $X$ is discrete, the probability can be estimated as a fraction of observations $\hat{p}(x)=\frac{\# x}{N}$, taking on value $x$ from the total $N$. An event $x \in \mathcal{X}$ is indicated as a draw of a value from the set. Also let $M$ be the total number of features in $X$, with $X_i,(i \in [1:M])$ as the current feature which is considered for selection, $y$ denote its label and $\Omega$ denote the sparse set of selected features with elements $x_{f_t}$ for $t \in [1,T]$, where $|\Omega|=T$ denotes its cardinality.

\subsection{Definitions}

Entropy and mutual information are two well-known concepts in information theory, which are used to measure the information provided by random variables.

\noindent \textbf{Entropy $H(X)$:} It is a measure of uncertainty of the random variable $X$, or the average amount of information that we receive with every event.  The more certain $X$ is, higher the value of $H(X)$, 
\begin{align*}
	H(X) &=  - \sum\limits_{x \in \mathcal{X}}P(X = x) \log P(X = x)\\
	&= - \sum\limits_{x \in \mathcal{X}}p(x) \log p(x)\numberthis \label{eq:entropy}
\end{align*}

\noindent \textbf{Joint entropy $H(X,Y)$:} It is the measure of uncertainty of a joint variable, which consists of two random variables $X$ and $Y$. The joint entropy of a set of variables is greater than or equal to all of the individual entropies of the variables in the set.
\begin{align*}
	H(X, Y ) = - \sum\limits_{x \in \mathcal{X},y\in \mathcal{Y}} p(x, y) \log (p(x, y))\numberthis \label{eq:jentropy}
\end{align*}
where $p(x,y) = P(X=x,Y=y)$.

\noindent \textbf{Conditional Entropy $H(X|Y)$:} It quantifies the amount of information needed to describe the outcome of a random variable, given that the value of another random variable.
\begin{align*}
	\notag H(X|Y ) &=- \sum\limits_{x \in \mathcal{X}}p(x) \sum\limits_{y \in \mathcal{Y}} p(y|x) \log (p(y|x)) \\
	&= -\sum\limits_{x\in \mathcal{X},y\in \mathcal{Y}} p(x, y) \log (p(x|y))\numberthis \label{eq:centropy} 
\end{align*}
where $p(x|y) = P(X = x|Y = y)$.

\noindent \textbf{Mutual Information (MI) $I(X:Y)$:} It is a measure of shared information between two random variables. MI between two r.v. $X$ and $Y$ can be defined as in Eq. \ref{eq:MI}:
\begin{align*}
	I(X:Y) &= H(X) - H(X|Y)\\ 
	&= H(X) + H(Y) - H(X,Y) \numberthis \label{eq:MI}
\end{align*}
which can be regarded as the certainty gain of random variable $X$ when conditioned of label $Y$. In this case, the higher value of $I$ means the stronger relation between $X$ and $Y$.

\subsection{Prior Work}

Lewis~\cite{lewis1992feature} and Duch~\cite{mim2006} coined the term Mutual Information Maximization (MIM) as a feature scoring criteria, which measures the usefulness of a feature subset when used for classification. This heuristic considers a score for each feature independently of others, which is a limitation by itself, yet is known to be suboptimal in those cases.
\begin{align*}
	J_{MIM}(x_i)=I(x_k:y)
\end{align*}
Battiti~\cite{battiti1994using} defined the Mutual Information Feature Selection (MIFS) criteria which ensures feature relevance and forces a penalty to ensure low correlations within the set of selected features, in a sequential manner. Setting the penalty term $\beta$ to zero would result in MIM scores.
\begin{align*}
	J_{MIFS}(x_i)=I(x_i:y)-\beta \sum_{x_j \in \Omega} I(x_i:x_j)
\end{align*}
Yang and Moody~\cite{yang1999data} and Meyer \etal~\cite{meyer2008information} proposed an alternate version of MIFS, called the Joint Mutual Information (JMI), which increases complementary information between features. This is able to capture the relevance-redundancy trade-off with various heuristic terms.
\begin{align*}
	J_{JMI}(x_i)= \sum_{x_j \in \Omega} I(x_ix_j:y)
\end{align*}
Peng \etal~\cite{peng2005feature} gave the Minimum-Redundancy Maximum-Relevance (MRMR) criteria which omits the conditional relevance term completely. The penalty term is inversely proportional to the size of current feature set. As the set grows the penalty term will diminish, hence all selected features will become strongly pairwise independent. 
\begin{align*}
	J_{MRMR}(x_i)= I(x_i:y) - \frac{1}{|\Omega|}\sum_{j \in \Omega} I(x_i:x_j)
\end{align*}
Fleuret~\cite{fleuret2004fast} introduced the Conditional Mutual Information Maximization (CMIM) criteria, which assumes that selected features are independent and class-conditionally independent given the unselected feature. This criterion ensures a good trade-off between independence and discrimination.
\begin{align*}
	J_{CMIM}(x_i)= I(x_i:y) - \underset{x_j \in \Omega}{max}[I(x_i:x_j)-I(x_i:x_j|y)]
\end{align*}
Nguyen \etal~\cite{nguyen2014effective} proposed a Global MI-based feature selection via spectral relaxation (SPEC\begin{tiny}CMI\end{tiny}) approach. They have the ability to handle second-order feature dependency, by favoring features having large total pairwise conditional relevance. They also show that for large data, low rank approximation can be applied to gain computational advantage to our global algorithm over its greedy counterpart.
\begin{align*}
	J_{SPEC_{CMI}}(x_i)= I(x_i:y) + \sum_{x_j \in \Omega} I(x_i:y|x_j)
\end{align*}
Gao \etal~\cite{gao2016variational} produced Variational Information Maximization (VMI) which can be applied for feature selection over any general class of distributions and provide tractable lower bounds for mutual information. Their model is optimal if the data is generated according to tree graphical models. They also outperform previous methods in speed. 
\begin{align*}
	J_{VMI}(x_i)= \argmax_{i \notin \Omega,x_j \in \Omega}I_{LB}(x_j \cup x_i:y)
\end{align*} 

\section{MI based Feature Selection} \label{sec3}

\subsection{Forward Heuristic}

Consider a supervised learning scenario where $x = \{x_1, x_2, ..., x_M\}$ is a $M$-dimensional input feature vector, and $y$ is the output label. In filter methods, the mutual information-based feature selection task is to select $T$ features $x_{\Omega^{T^*}} = \{x_{f_1}, x_{f_2}, ..., x_{f_T} \} $ such that the mutual information between $x_{\Omega^*}$ and $y$ is maximized. Formally,
\begin{align}
	\Omega^* = \arg\max\limits_{\Omega}I(\Omega:y) \qquad s.t. \quad |\Omega| = T \label{eq:forwardS}
\end{align}
where $I(\cdot)$ denotes the mutual information.

Directly optimizing Eq.~\ref{eq:forwardS} is a NP-Hard combinatorial problem~\cite{amaldi1998approximability}. Thus, most of MI-based features selection methods use a greedy search strategy, which selects features incrementally. 
\begin{align}
	f_t = \arg\max\limits_{i}\{ I(\Omega^{t-1}\cup x_i: y)\}, i \in [1,M] \label{eq:ft}
\end{align}
Because we just focus on the next feature $x_i$ to be considered for selection, this becomes equivalent to solving Eq.\ref{eq:ftfinal}. Readers can refer to \cite{gao2016variational} for a detailed derivation.
\begin{align}
	f_t = \arg\max\{(I(x_i:y) + H(x_{\Omega^{t-1}}|x_i) - H(x_{\Omega^{t-1}}|x_i,y)\} \label{eq:ftfinal}
\end{align}

\subsection{Lower Order Approximation}

Approximating the higher order information-theoretic  based MI measure is a difficult task. Therefore, most current methods propose a relevant term (features having information which is already contained in other features) and a redundant term (features with no useful information related to the class variable), where Feature Independence and Class-conditioned Independence are assumed (e.g. Eq.~\ref{eq:assm1} and ~\ref{eq:assm2}).
\begin{align}
	H(x_{\Omega^{t-1}}|x_i) &\approx \sum\limits_{k=1}^{t-1}H(x_{f_k}|x_i)\label{eq:assm1}\\ 
	H(x_{\Omega^{t-1}}|x_i,y) &\approx \sum\limits_{k=1}^{t-1}H(x_{f_k}|x_i,y)\label{eq:assm2}
\end{align}

Gao \etal~\cite{gao2016variational} discusses the inconsistency of these two assumptions. The independence assumption leads to the miss of multi-way feature combinations, because the interdependence between features is ignored. 

Let us look at a scenario where a lower order MI based feature selection algorithm would fail. For example, consider three random variables $x_1,x_2,x_3$, where $x_1,x_2$ have the same pdf: $p(x=0)=0.5$ and $p(x=1)=0.5$, and $p(x_3=0)=1$. Let $Y=x_1 \odot x_2$. In that case, $I(x_1 \cup x_2:y) = 1$, however, $I(x_1:y)+I(x_2:y)=0$. Thus a lower order MI based algorithm should select $x_3$, which is irrelevant to $y$ and ignore the feature combination $x_1 \cup x_2$, which fully defines $y$. 

\subsection{Independent Component Analysis}

The starting point for ICA is the very simple assumption that input variables $x_i$ can be represented as a linear mixture of $N$ independent components $s_1,s_2,\ldots,s_N$.
\begin{align*}
	x_i = \sum\limits_{j=1}^{T} \alpha_{ij} s_j  \qquad i = 1,2,\ldots,T
\end{align*}
When writing into a compact version, we have
\begin{align}
	\notag X =AS \quad \textrm{or} \quad S=WX, \quad \textrm{where} \quad W = A^{-1}
\end{align}
Based on maximum likelihood estimation, ICA algorithm constructs unmixing matrix $W$ that gives the best fit of input data vectors $x_i$ given joint pdf of signals vectors.

In this context, both $W$ and $S$ are unknown, and we further assume that signal variables' cumulative distribution function (cdf) fit a sigmoid function.
\begin{align*}
	g(s) = \frac{1}{1+e^{-s}}
\end{align*}
Thus, the unmixing matrix $W$ is defined as follow in Eq.~\ref{eq:defW}, and the log-likelihood function can be optimized by stochastic gradient descend using Eq.~\ref{eq:iterW}. 
\begin{align}
	W = \arg\max\{\sum\limits_{i=1}^{M}(\sum\limits_{j=1}^{T}log(g'(w_j^Tx^{(i)})+log(|W|))\} \label{eq:defW}
\end{align}
where each $x^{(i)}$ denotes a sample of $T$ selected features.
\begin{align}
	W := W + \alpha(1-2g(Wx_{batch})x^{T}_{batch} + (W^T)^{-1})\label{eq:iterW}
\end{align}
where $x_{batch}=[x^{(1)},x^{(2)},\ldots ,x^{(m)}]$ and $\alpha $ is learning rate.

\section{Proposed Method: Higher Order Feature Selection} \label{sec4}

We are now considering finding out inter-dependent feature combinations, which will be missed by only implementing lower order MI approximation. This would jointly address the problem of feature selection and feature ranking \footnote{Our code shall soon be made available online 
from \href{https://cvcweb.ices.utexas.edu/cvcwp/software/}{https://cvcweb.ices.utexas.edu/cvcwp/software/}}.

\subsection{Motivation}

The problem of approximating joint mutual information(MI) as the summation of single variables is limited by the inter-dependence between selected features. That motivates us to determine $K$ independent feature subsets during the greedy search (Eq.~\ref{eq:ft}) process $X_{\Omega^{1}},X_{\Omega^{2}},\ldots, X_{\Omega^{K}}$, where features within a subset can be dependent on each other. Secondly, we determine independent representation $S_{\Omega^{1}},S_{\Omega^{2}},\ldots,S_{\Omega^{K}}$ for each of the independent feature subsets, then reconstruct $H( X_{\Omega^{i}})$ using $H( S_{\Omega^{i}})$. Potential problems of this framework lies  in the accuracy of recovering $H( X_{\Omega^{i}})$ using $H(S_{\Omega^{i}})$ and the computational cost of ICA algorithm. We solve both of them by the following mutual balance and incremental ICA algorithm.

\subsection{Subset Independence}

Instead of assuming single feature independence and class-conditioned independence, we assume that the optimal selected feature subset can be divided into several feature subsets, which are independent and Class-Conditioned Independent (Eq.~\ref{eq:newassm1} and ~\ref{eq:newassm2}).

The rationality of the subset assumption is that in most cases, label $y$ can be better approximated as some function of $X_{\Omega^i}$, e.g.
\begin{align}
	y = f(\Omega) = f(X_{\Omega^1},X_{\Omega^2},…,X_{\Omega^{K}})
\end{align}
where $K$ indicates the number of selected subsets.

Let $\Omega = \bigcup\limits_{i=1}^K X_{\Omega^i}$ ,where each $X_{\Omega^i}$ denote a feature subset and $x_t$ denotes the next feature to be selected.
\begin{align}
	Assumption 1.\qquad &P(X) = \prod\limits_{i=1}^K P(X_{\Omega^i}) \label{eq:newassm1}\\
	Assumption 2.\qquad &P(X|y) = \prod\limits_{i=1}^K P(X_{\Omega^i}|y) \label{eq:newassm2}
\end{align}
\begin{figure}[ht]
	\centering
	\begin{subfigure}{0.49 \textwidth}
		\centering
		\includegraphics[scale=0.5]{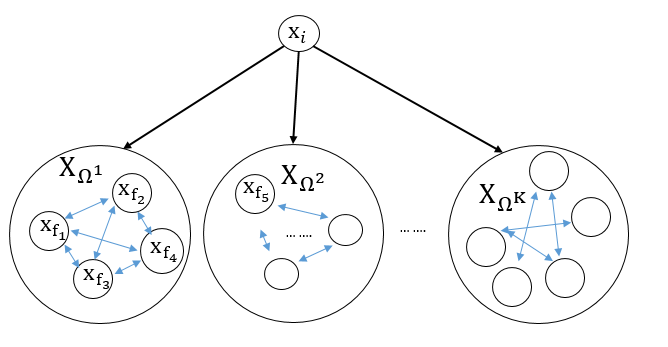}
		\caption{Assumption 1: Feature Subset Independence}
	\end{subfigure}\\
	\begin{subfigure}{0.49 \textwidth}
		\centering
		\includegraphics[scale=0.5]{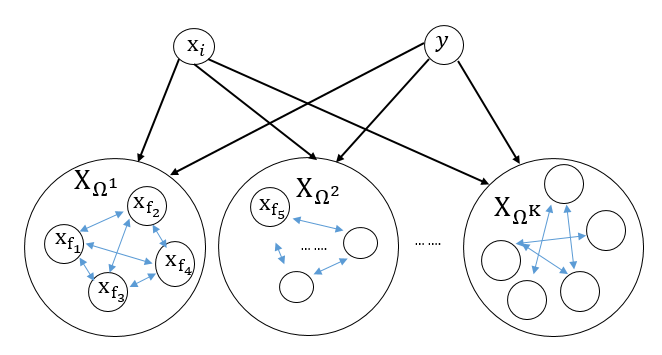}
		\caption{Assumption 2: lass-Conditioned Subset Independence}
	\end{subfigure}
	\caption{New Assumptions for our model HOFS. Blue arrows show the dependence between selected features in the same subset while subsets are independent of each other.}
	\label{fig:assump}
\end{figure}

\subsection{Entropy relation between Inputs and Signals}

Joint entropy of feature combinations $X_{\Omega^{i}}$ is computed using Eq.~\ref{eq:jentropyHS}\\

\begin{align}
	H(X_{\Omega^i}) = -\sum\limits_{x \in X_{\Omega^i}} p(X_{\Omega^i})log(p(X_{\Omega^i})) \label{eq:jentropyHS}
\end{align}

However, the joint pdf of $X_i$ is difficult to estimate using finite samples as the size of subset increases. Directly computing that is also an a NP-Hard combinatorial problem which leads to inaccuracy because of insufficient sampling.

To solve that, we first transform the original feature vectors $X_{\Omega^i}$ to $S_{\Omega^i}$ using ICA. Note that each feature vector in $S_{\Omega^i}$ is independent with each other. Thus $H(X_{\Omega^i})$ can be computed using $H(S_{\Omega^i})$, which is a summation of $H(s_j)$, where $s_j \in S_{\Omega^i}$, shown in Eq.~\ref{eq:proofHS}.
\begin{align*}
	H(S_{\Omega^i}) &=- \sum\limits_{s_1,...,s_{T_i}} p(s_1,... ,s_{T_i})log(p(s_1,... ,s_{T_i}))\\
	&= -\sum\limits_{s_1}\sum\limits_{s_2}\dots\sum\limits_{s_{T_i}}\prod\limits_{j=1}^{T_i} p(s_j)(\sum\limits_{j=1}^{T_i}log(p(s_i)))\\
	&= - \sum\limits_{j=1}^{T_i}(\sum\limits_{s_1}\sum\limits_{s_1}\dots \sum\limits_{s_{T_i}}(\prod\limits_{j=1}^{T_i} p(s_j))log(p(s_j)))\\
	&=-\sum\limits_{j=1}^{T_i} \sum\limits_{s_j} p(s_i)log(p(s_i))\\
	&= \sum\limits_{j=1}^{T_i} H(s_j) \numberthis\label{eq:proofHS}
\end{align*}
Noticing the relation of pdf between X and S in Eq.~\ref{eq:ralationF} and ~\ref{eq:ralationP}, $H(X_\Omega)$ can be computed using Eq.~\ref{eq:ralationHS} \\
\begin{align}
	\notag F_X(x) &= P(X\le x) = P(AS \le x)\\
	&= P(S\le Wx) = P(S \le s)= F_S(s) \label{eq:ralationF}
\end{align}  
\begin{align}
	P_X(x) &= \nabla_X F_X(x) = \nabla _S F_S(s) * \nabla_X S = P_S(s)|W|\label{eq:ralationP}
\end{align}
Thus, $H(X_{\Omega})$ can be formulated as:
\begin{align*}
	H(X_{\Omega}) &= -\sum\limits_x p_X(x)log(p_X(x))\\
	&= -\sum\limits_x P_X(x)log(P_S(s)|W|)\\
	&= -\sum\limits_x P_X(x)log(P_S(s)) -log(|W|) \\
	&= -\sum\limits_s P_S(s)log(P_S(s)) -log(|W|) \\
	&= H(S_{\Omega}) - log(|W|)\\
	&= \sum\limits_{i=1}^MH(s_i) - log(|W|) \numberthis \label{eq:ralationHS}
\end{align*}

\subsection{Forward Search}

Based on the Assumption 1 and 2 (Eq.~\ref{eq:newassm1} and Eq.~\ref{eq:newassm2}), the forward heuristic search function Eq.~\ref{eq:ftfinal} becomes:
\begin{align}
	\notag f_t &= \arg\max\limits_{i \notin \Omega}I(x_{i}:y) + H(\Omega|x_i) - H(\Omega|x_i,y)\\
	&=  \arg\max\limits_{i \notin \Omega}I(x_{i}:y) + \sum\limits_{j=1}^K (H(X_{\Omega^{j}}|x_i)-H(X_{\Omega^{j}}|x_i,y)) \label{eq:forwsearch}
\end{align}
The computation $H(X_{\Omega_{j}^{t-1}}|x_i)$ and $H(X_{\Omega_{j}^{t-1}}|x_i,y)$ is still a NP-Hard problem and finite discrete samples of input vectors $X$ deteriorate the approximation of joint pdf. To solve this problem, we firstly transform them into non-conditioned terms as Eq.~\ref{eq:uncondH1},~\ref{eq:uncondHy}. By assuming some general cumulative distribution function of signal vectors $s_i$, we calculate those two terms in linear computational complexity using Eq.~\ref{eq:forwsearch}, thus avoiding computation of the joint pdf.
\begin{align}
	&H(X_{\Omega^{t-1}}|x_i) = - H(x_i) +  H(X_{\Omega^{t-1}}\cup x_i) \label{eq:uncondH1}\\
	&H(X_{\Omega^{t-1}}|x_i,y) = -H(x_i,y) +  H(X_{\Omega^{t-1}}\cup x_i\cup y)\label{eq:uncondHy}
\end{align}

\subsection{Inaccuracy Removal and Incremental ICA}

The transformation from input vectors $X_{\Omega^i}$ to independent signal vectors $X_{\Omega^i}$ assumes continuous distribution. Finite discrete samples lead to inaccuracy.
\begin{align}
	H(X_{discrete}) \approx H(S_{discrete}) - log(|W|)
\end{align}
Note that we try to find a mixture of independent signals $s_i$ to represent $x_i$. The inaccuracy comes from $H(X_{{\Omega}^{t-1}}|x_i)-H(X_{{\Omega}^{t-1}}|x_i,y) $ in Eq.~\ref{eq:forwsearch}. Since forward greedy search concern a subtraction of two multi-variables entropies we are motivated to design a balanced  algorithm to reduce noise. The computational cost for implement ICA algorithm is not tractable either. Hence, we adopt another assumption that the new input vector can be represented as a linear combination of our origin signal set $S_{S^j}$ and a new signal vector $s_t$. That is, given $X_{{\Omega}^{j}} = AS_{{\Omega}^{j}} $, where we can obtain another independent signal vector $s_t$ make the following equation hold true.
\begin{align*}
	\begin{pmatrix} X_{\Omega^{j}} \\  x_i  \end{pmatrix} = 
	\begin{pmatrix} A_{\Omega^{j}} & 0 \\ a_{T_i,1},...,a_{T_i,T_i-1} & a_{T_i,T_i}  \end{pmatrix}
	\begin{pmatrix} S_{\Omega^{j}} \\  s_t  \end{pmatrix} =  A_{S{j}}S_{\Omega^t}
\end{align*}
Thus,
\begin{align*}
	W_{\Omega^j}^i &= (A_{\Omega^j}^i)^{-1}\\ &= 
	\begin{pmatrix} A^{-1}_{\Omega^j} & 0 \\w_{T_i+1,1},\dots,w_{T_i+1,T_i} & a^{-1}_{T_i+1,T_i+1} \end{pmatrix}\\&= 
	\begin{pmatrix} W_{\Omega^j} &0 \\  w_{T_i+1,1},\dots,w_{T_i+1,T_i} & w_{T_i+1,T_i+1} \end{pmatrix}
\end{align*}
In this context, the ICA algorithm can be done incrementally and reduce the computational cost from computing the gradient of whole matrix $A$ to only the last row of $A$. And because $A$ is a lower triangular matrix, computational complexity drops from $O(T^3 )$ to $O(T^2 )$.

Given $X_{\Omega^{j}}= A_{\Omega^{j}}S_{\Omega^{j}}$, we imply incremental ICA for the next feature $x_i$ and its label $y$.
\begin{align*}
	\begin{pmatrix} X_{\Omega^{j}} \\  x_i  \end{pmatrix} &= 
	\begin{pmatrix} A_{\Omega^{j}} & 0 \\ a^i_{T_j+1,1},...,a^i_{T_j+1,T_j} & a^i_{T_j+1,T_j+1}  \end{pmatrix}
	\begin{pmatrix} S_{\Omega^{j}} \\  s_{x_i}  \end{pmatrix} \\
	&=  A_{x_i}S'_{\Omega^j}
\end{align*}
and
\begin{align*}
	\begin{pmatrix} X_{\Omega^{j}} \\  x_i \\y \end{pmatrix} = 
	\begin{pmatrix} A_{x_t} & 0 \\ a^i_{T_j+2,1},...,a^i_{T_j+2,T_j+1} & a^i_{T_j+2,T_j+2}   \end{pmatrix}
	\begin{pmatrix} S_{\Omega^{j}} \\  s_{x_i}\\s^i_y \end{pmatrix}
\end{align*}
So, the subtraction of $H(X_{\Omega^{j}}|x_i)-H(X_{\Omega^{j}}|x_i,y)$ can be computed as Eq.~\ref{eq:IICA}. 
\begin{align*}
	& H(X_{\Omega^{j}}|x_i)-H(X_{\Omega^{j}}|x_i,y) \\
	=& - H(x_i) + H(x_i,y) - H(X_{\Omega^{j}}\cup x_i)+H(X_{\Omega^{j}}\cup x_i\cup y)\\
	=& - H(x_i) + H(x_i,y) \\
	&- (\sum\limits_{i=1}^{T_j}H(s_i) + H(s_{x_i}) + log(|A_{x_i}|)) \\
	&+ (\sum\limits_{i=1}^{T_j}H(s_i) + H(s_{x_i}) + H(s^i_y)  + log(|A_{x_i}|a^i_{T_j+2,T_j+2}))\\
	=&- H(x_i) + H(x_i,y) + H(s^i_y) + log(a^i_{T_j+2,T_j+2}) \numberthis \label{eq:IICA}
\end{align*}\\
Let $I(X_{\Omega^j}:y|x_i)$ denote $H(X_{\Omega^{j}}|x_i)-H(X_{\Omega^{j}}|x_i,y)$\\
The forward search function becomes,
\begin{align}
	f_t= \arg\max\limits_{i \notin \Omega^{t-1}}\{I(x_{i}:y) + \sum\limits_{j=1}^K( I(X_{\Omega^j}:y|x_i))\} \label{eq:finalforward}
\end{align}

\subsection{Time Complexity}

A detailed stepwise description of the HOFS algorithm is given in Alg.~\ref{alg:HOFS}. Because the forward heuristic search at each step will compute the information gain between each unselected node and selected node, the total computational complexity for entropy is $O(MNT)$, where $N$ is the total number of samples, $M$ is the total number of features and T is the number of features to be selected.

The complexity of ICA algorithm concerns the computation of gradients of matrix $W_{\Omega^i}$ (Eq.~\ref{eq:iterW}). Because $W_{\Omega^i}$ is lower triangular matrix and we only need the gradients of elements in the last row, Inverse computation can be done in $O(\frac{MT^2}{K})$.

\begin{algorithm}[h]
	\caption{Higher Order Feature Selection (HOFS)}
	\begin{algorithmic}[1]
		\BState \emph{\textbf{Data}}:$(X_i,y_i) \in \mathcal{R}^N, i=1,2,..,M$
		\BState \emph{\textbf{Input}}:$T \leftarrow $ number of features to be selected.
		\State $\Omega \leftarrow \{ \emptyset \}, K \leftarrow 1, t\leftarrow1 $
		\BState \emph{\textbf{Body}}:
		\While{$|\Omega|<T$}
		\State imply ICA for each unselected feature $x_i$
		\State select a new feature $x_{f_t}$ using Eq.~\ref{eq:finalforward}
		\State compute $Argcov$ and $Maxcov$ using Eq.~\ref{eq:acov} and Eq.~\ref{eq:mcov}
		\If {$ Maxcov > C $}
		\State  $K \leftarrow \argmax Maxcov$
		\Else 
		\State  $K \leftarrow |\Omega|$
		\EndIf
		\State  $X_{\Omega^{K}} \leftarrow X_{\Omega^{K}} \cup x_{f_t}  $
		\State  $\Omega \leftarrow \Omega \cup x_{f_t}  $
		\EndWhile
		\BState \emph{\textbf{Output}}:
		\State \textit{selected feature subsets:} $\Omega$ 
	\end{algorithmic}
\end{algorithm}

\subsection{Feature Subset Determination}

We aim to find the independent feature combinations, so it is of utmost importance to judge if a new subset should be created for incoming feature $x_t$ or put it into an existing subset. Specifically, we compute the maximum average correlation between $x_t$ and each subset $X_{{\Omega}^i}$. 
\begin{align}
	Argcov_{i} =  \frac{1}{|X_{{\Omega}^i}|}\sum\limits_{u \in X_{S^i}}cov(x_t,u) \label{eq:acov}\\
	Maxcov = \max\limits_{i} Argcov \label{eq:mcov}
\end{align}
If $Maxcov$ is over a predefined constant $C$, then we add $x_t$ to the relevant subset, else we create a new subset for $x_t$.

\section{Experiments} \label{sec5}

We conduct experiments by fisrt analyzing the quality of the incremental ICA algorithm. We then compare our proposed approach HOFS, with other popular lower order MI based feature selection algorithm on some publicly available standard datasets.

\subsection{Incremental ICA Quality}

We verify the feasibility of our incremental ICA algorithm by two measures (Table~\ref{table:pearson}). Firstly, we compute the average Pearson's product-moment coefficient between the signal vectors after ICA, to show that by maximizing Eq.~\ref{eq:defW}, we get uncorrelated vectors, which ensures that Eq.~\ref{eq:proofHS} holds true. Low values $(\sim 0.055)$ on average, show that the vectors in the ICA space are highly uncorrelated to each other, hence almost independent. Secondly, we compute average $R_{balance}$ ratio:
\begin{align*}
	R_{balance}= \frac{1}{K}\sum\limits_{i = 1}^K \frac{H(X_{\Omega^i})-H(X_{\Omega^i},y)}{H(s^i_y) + log(a^i_{T_i+2,T_i+2})}
\end{align*}
where $X_{\Omega^i},i = 1,2,3,...K$ is the $i^{th}$ selected feature combination, $s^i_y$ is the incremental signal vector for $y$ in subset $X_{\Omega^i}$. With $R_{balance}$ near to $1$, we show that the error of approximating $H(X)$ using $H(S)$ is balanced out by the subtraction of $H(X)-H(X,y)$.

\begin{table*}[h]
	\centering
	\begin{tabular}{|c|c|c|c|c|c|} 
		\hline
		Dataset & Features & Samples & Classes & Avg. Pearson Coeff. & Avg. $R_{balance}$\\\hline
		Lung  \cite{ding2005minimum} & 56 & 32 & 3 & 0.01 & 1.1\\
		Splice \cite{bache2013uci}   & 60 & 3175 & 3 & 0.05 & 1.07\\
		Waveform \cite{bache2013uci} & 40 & 5000 & 3 & 0.04 & 0.94\\
		Semeion \cite{bache2013uci}  & 256 & 1593 & 10 & 0.09 & 0.94\\
		Optdigits\cite{bache2013uci} & 64 & 5620 & 10 & 0.11 & 0.89\\
		Musk2   \cite{bache2013uci}  & 168 & 6598 & 2 & 0.07 & 0.98\\
		Spambase\cite{bache2013uci}  & 57 & 4601 & 2 & 0.06 & 0.97\\
		Promoter \cite{bache2013uci} & 58 & 106 & 4 & 0.04 & 1.02\\
		Madelon \cite{guyon2003introduction} & 500 & 4400 & 2 & 0.02 & 1.004\\\hline
	\end{tabular}
	\caption{Incremental ICA Quality. Low values of average Pearson Coefficient shows low correlation among the feature subsets after ICA algorithm. Average $R_{balance}$, shows our approximation ratio using entropy from the ICA space.}
	\label{table:pearson}
\end{table*}

\subsection{Real-World Data}

We compare our algorithm with other popular information-theoretic feature selection methods, including VMI, mRMR, JMI, CMIM and SPEC\begin{tiny}CMI\end{tiny}. We use $9$ well-known datasets commonly used in feature selection studies. They were chosen to have a wide variety of feature ratios and multi-class problems (Table~\ref{table:pearson}). We use the average cross-validation error rate on the range of $10$ to $100$ (total feature number if it is less than $100$) features to compare different algorithms under the same setting. $10$-fold cross-validation is employed for datasets with number of samples $N \geq 100$ and leave-one-out cross-validation otherwise. The classifier is chosen to be Linear SVM. We pre-process data following the approach proposed in VMI~\cite{gao2016variational}.
\begin{figure}[h]
	\centering
	\includegraphics[width=0.5\textwidth]{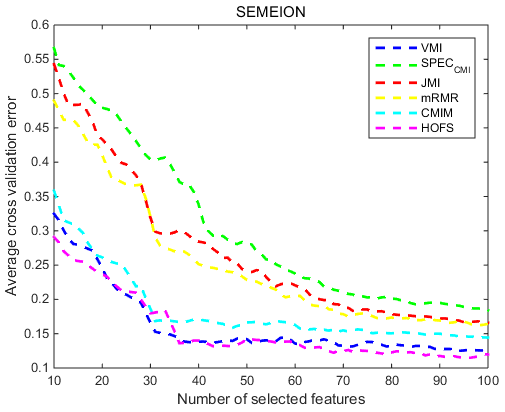}
	\caption{Average Cross Validation Error of the aforementioned methods and ours, on the SEMEION~\cite{bache2013uci} dataset. It consists of $256$ features, $1593$ samples and $10$ classes. $10$-fold cross-validation using a Linear SVM is employed for this dataset.}
	\label{fig:crossvalerr}
\end{figure}

Fig.~\ref{fig:crossvalerr} shows the cross validation error in dataset `Semeion', where our method incurs the lowest validation error in terms of classification using the selected feature sets. Highlighted entries in each table shows the best results obtained during our experimentation. We show the average cross validation error results for the nine datasets~\cite{bache2013uci} in Table~\ref{table:crossval}. Our method performs best for most of the datasets and has very low relative error compared to others, when it is not the best. 
\begin{table*}[h]
	\centering
	\begin{tabular}{|c|ccccc|c|}
		\hline
		Dataset & VMI\cite{gao2016variational} & mRMR\cite{peng2005feature} & JMI\cite{yang1999data} & CMIM\cite{fleuret2004fast}& SPEC\begin{tiny}CMI\end{tiny}\cite{nguyen2014effective} & Our Method\\
		\hline
		Lung      & 7.4 & 10.9 & 11.6 & 11.4 & 11.6 & \textbf{6.9}\\
		Splice    & 13.7 & 13.6 & 13.7 & 14.7 & 13.7 & \textbf{13.2}\\
		Waveform  & \textbf{15.9} & \textbf{15.9} & \textbf{15.9} & 16.0 & \textbf{15.9} & 16.0\\
		Semeion   & 14.0 & 23.4 & 24.8 & 16.3 & 26.0 & \textbf{13.7}\\
		Optdigits & \textbf{7.2} & 7.6 & 7.6 & 7.5 & 9.2 & 7.4\\
		Musk2     & 12.6 & 12.4 & 12.8 & 13.0 & 15.1 & \textbf{11.9}\\
		Spambase  & \textbf{6.6} & 6.9 & 7.0 & 6.8 & 9.0 & \textbf{6.6}\\
		Promoter  & \textbf{21.2} & 21.5 & 22.4 & 22.1 & 24.0 & \textbf{21.2}\\
		Madelon   & 16.7 & 30.8 & \textbf{15.3} & 17.4 & 15.9 & 16.0\\\hline
	\end{tabular}
	\caption{Average $10$-fold Cross Validation Error Rate Comparison using a Linear SVM. Lower values are better.}
	\label{table:crossval}
\end{table*}

The global Mutual Information captured by the various methods for top $11$ features is shown in Table~\ref{table:globalmi}, where our method returns relatively high MI compared to others by choosing the best possible combination of features. This further goes to ground our assumption that certain set of features when selected together, rather than independently, have a higher impact on global MI.
\begin{table*}[h]
	\centering
	\begin{tabular}{|c|ccccc|c|}
		\hline
		Dataset & VMI\cite{gao2016variational} & mRMR\cite{peng2005feature} & JMI\cite{yang1999data} & CMIM\cite{fleuret2004fast}& SPEC\begin{tiny}CMI\end{tiny}\cite{nguyen2014effective} & Our Method\\
		\hline
		Lung      & 82.342 & 82.973 & 82.954 &	82.909 & 82.878 &	\textbf{83.909}\\
		Splice    & 10.462 & 10.469 & \textbf{10.565} & 10.447 & 10.447 & 10.471\\
		Waveform  & 46.231 & 36.616 & 11.607 &	54.021 & 54.021 & \textbf{59.035}\\
		Semeion   & 5.231 & 4.328 & 4.788 & 3.742 &	3.556 & \textbf{5.556}\\
		Optdigits & 157.239 & 141.055 & 168.096 & 141.055 & 141.059 & \textbf{171.061}\\
		Musk2     & 11.675 & 2.612 & 10.006 & 3.705 & 2.865 & \textbf{13.412}\\
		Spambase  & 5.689 & 5.204 & 5.662 & 5.204 & 5.151 & \textbf{5.950}\\
		Promoter  & 11.410 & 11.410 & 11.411 & 11.380 & 11.398 & \textbf{11.411}\\
		Madelon   & 14.284 & 1.797 & 7.099 & 4.347 & 6.084 & \textbf{18.009}\\\hline
	\end{tabular}
	\caption{Global Mutual Information for the top $11$ features. Higher values are better.}
	\label{table:globalmi}
\end{table*}

We also consider the Average Relative Absolute Error $(ARAE)$ of the classification process (Table~\ref{table:arae}). This parameter shows how a feature selection method could affect the classifiers not to predict wrongly or at least predict closer to the true labels. Considering $RAE_i$ as the relative absolute error for a specific classification algorithm, and $Q$ as the number of such algorithms used in the experiment, ARAE is then defined as:
\begin{align*}
ARAE = \frac{\sum_{i=1}^{Q}RAE_i}{Q}
\end{align*}
\begin{table*}[h]
	\centering
	\begin{tabular}{|c|ccccc|c|}
		\hline
		Dataset & VMI\cite{gao2016variational} & mRMR\cite{peng2005feature} & JMI\cite{yang1999data} & CMIM\cite{fleuret2004fast}& SPEC\begin{tiny}CMI\end{tiny}\cite{nguyen2014effective} & Our Method\\
		\hline
		Lung      & 23.1 & 28.4 & 31.2 & 24.3 &	15.2 & \textbf{11.4}\\
		Splice    & 20.9 & 19.6 & 20.2 & 21.3 &	23.9 & \textbf{19.2}\\
		Waveform  & 15.5 & 16.0 & 15.1 & 15.3 &	16.2 & \textbf{15.0}\\
		Semeion   & \textbf{16.2} & 28.1 & 30.2 & 18.8 &	39.2 & \textbf{16.2}\\
		Optdigits & \textbf{9.2} & 11.7 & 10.3 &	16.0 & 14.1 & \textbf{9.2}\\
		Musk2     & 14.2 & 17.2 & 15.9 & 17.3 &	17.3 & \textbf{12.0}\\
		Spambase  & \textbf{7.8} & 8.8 & 14.2 & 13.7 & 19.1 & \textbf{7.8}\\
		Promoter  & 31.7 & 31.7 & \textbf{29.7} & 30.9 & 34.2 & 31.7\\
		Madelon   & 28.9 & 46.8 & 21.3 & 27.1 &	22.2 & \textbf{20.0}\\\hline
	\end{tabular}
	\caption{Average Relative Absolute Error $(ARAE)$ for top $30$ features using Linear SVM. Lower values are better.}
	\label{table:arae}
\end{table*}

\section{Synthetic Model Experiment} \label{sec6}

\subsection{Inference from Graphical Model}

\begin{figure*}[ht]
	\centering
	\begin{subfigure}{0.33 \textwidth}
		\centering
		\includegraphics[scale=0.5]{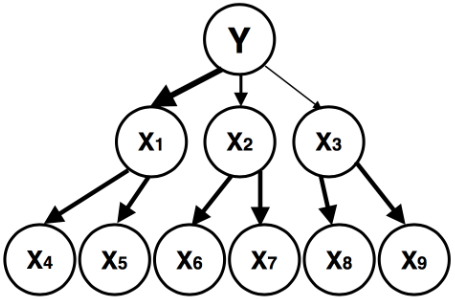}
		\caption{Model for synthetic experiment} \label{fig:model}
	\end{subfigure}
	\begin{subfigure}{0.33 \textwidth} 
		\centering
		\includegraphics[scale=0.25]{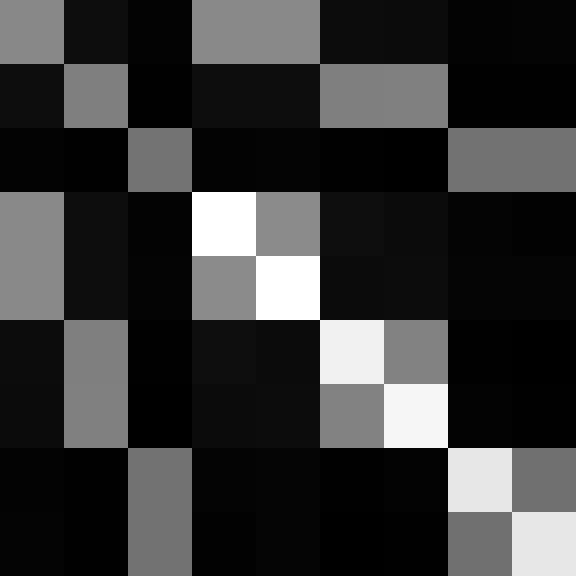}
		\caption{Covariance Matrix for the $9$ features} \label{fig:corr}
	\end{subfigure}
	\begin{subfigure}{0.33 \textwidth} 
		\centering
		\includegraphics[scale=0.4]{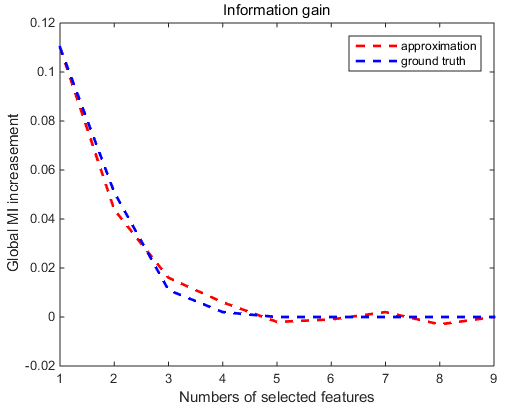}
		\caption{Global MI gain} \label{fig:globalMI}
	\end{subfigure}
	\caption{Generative Model for $N=100000$. (a) $Y$ is a binary label, while $x_i$ is continuous gaussian with unit variance and mean set to the value of its parent. Edge thickness represents the relationship strength. (b) Brighter area indicates larger covariance value, showing that those features are highly correlated. (c) Comparison between Ground truth MI and that computed by our approach. Features selected at each step in shown in Table~\ref{table:comp1}, in decreasing order of MI.}
	\label{fig:gen}
\end{figure*}

We perform experiment on a synthetic model (Fig.~\ref{fig:gen}) according to the tree structure. The root node $Y$ in Fig.~\ref{fig:model} is a binary variable indicating the class label, while other variables $x_i$ are continuous gaussian with unit variance and mean set to the value of its parent. We generate $100000$ samples from the model and compare the results of HOFS with VMI in Table~\ref{table:comp1}]. The covariance structure among the $9$ features (Fig.~\ref{fig:corr}) confirms the relation generated by the tree graphical model. Although both the methods selects the same set of features, HOFS selects them in the correct order and maintains the proper subsets. $\Omega_{VMI}=\{x_1,x_4,x_5,x_2,x_6,x_7,x_3,x_8,x_9\}$ and $\Omega_{HOFS}=\{\{x_1,x_4,x_5\},\{x_2,x_6,x_7\},\{x_3,x_8,x_9\}\}$. Therefore, HOFS goes an extra step in showing which features are correlated to each other, which is not evident from VMI. Fig.~\ref{fig:globalMI} shows the plot of Information gain according to the generated model and as predicted by HOFS. We also show the Information gain by adding a new feature to the current selected subset in Table~\ref{table:comp2}.  

\begin{table*}[h]
	\centering
	\begin{tabular}{|c|c|c|c|c|c|c|c|c|c|}
		\hline
		Feature & $x_1$ & $x_2$& $x_3$& $x_4$& $x_5$& $x_6$& $x_7$& $x_8$& $x_9$\\\hline
		$I(x_i:Y)$ VMI & \textbf{0.111} & 0.052 & 0.022 & \textbf{0.058} & \textbf{0.058} & 0.025 & 0.029 & 0.012 & 0.013\\\hline
		$I(x_i:Y)$ HOFS & \textbf{0.111} & 0.052 & 0.022 & \textbf{0.058} & \textbf{0.058} & 0.025 & 0.029 & 0.012 & 0.011\\\hline
	\end{tabular}
	\caption{Mutual Information for each feature on Synthetic Model (Fig.~\ref{fig:gen}). Top $3$ features are highlighted. This is evident from the model itself as feature $x_1$ has the highest correlation to label $y$ and features $x_4,x_5$ are correlated to $x1$.}
	\label{table:comp1}
\end{table*}
\begin{table*}[h]
	\centering
	\begin{tabular}{|c|c|c|c|c|c|c|c|c|c|}
		\hline
		Feature & $x_1$ & $x_2$& $x_3$& $x_4$& $x_5$& $x_6$& $x_7$& $x_8$& $x_9$\\\hline
		VMI & 0.111 & 0.0438 & 0.012 & 0.008 & 0.005 & 0.001 & -0.001 & 0.002 & -0.003\\\hline
		HOFS & 0.110 & 0.0438 &	0.016 &	0.006 &	-0.002 &	-0.001 & 0.002 & -0.003 & 0.00\\\hline
	\end{tabular}
	\caption{Information gain by adding a new feature on the Synthetic Model}
	\label{table:comp2}
\end{table*}

\subsection{Handling Non-Numerical Features}

To verify the effectiveness of the proposed HOFS for a non-/numerical feature selection, $1000$ samples with $20$ heterogeneous features and $5$ classes were synthesized according to~\cite{wei2015heterogeneous}. There were $4$ groups of features and each group had $5$ features of both numerical and non-numerical features, which are shown in Table~\ref{table:synth}. The class label can be described by the group of features. In Group I, $F_1,F_2,F_3$ are non-numerical features which can explain class 1 - class 3. $F_4,F_5$ are numerical features which can explain class 4 and class 5. Here, $F_1$ is the most prominent feature to classify the class label and there are redundancies within this group. Ideally, the selected feature subset should be $\{F_1,F_3,F_5\}$ or $\{F_1,F_4,F_5\}$. If anyone of these two subsets is selected, other features in this group are redundant. Group II contains $5$ features which were generated by adding some bias to the related features in Group I. Group III contains $5$ features which were generated by adding noise to the related features in Group I. Group IV contains $5$ random attributes, which were uncorrelated with class label.

\begin{figure}[h]
	\centering
	\includegraphics[scale=0.47]{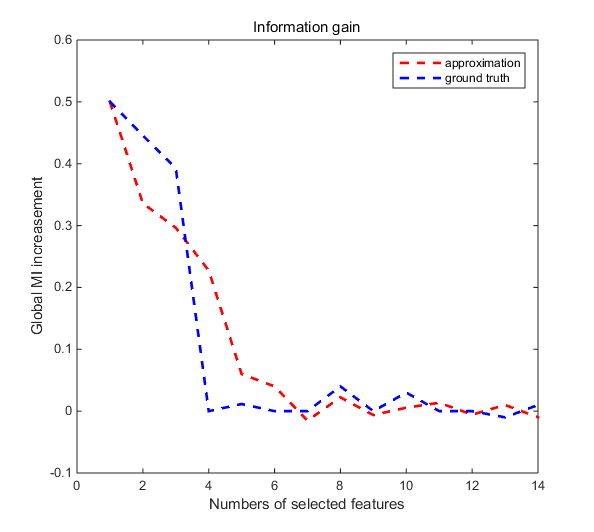}
	\caption{Global MI gain for top $14$ selected features. Last $6$ are not considered due to redundancy.}
	\label{fig:hetero}
\end{figure}

VMI completely breaks down in this scenario and selects the features in linear order as $\Omega_{VMI}=\{F_1,F_2,\ldots,F_{20}\}$. The subsets produced are $\Omega_{HOFS}=\{\{F_1,F_6,F_2,F_7\},\{F_9,F_4,F_{14},F_{10},F_5,F_{15}\},\{F_3,F_8\},\{F_{11}\},\{F_{13}\},\{F_{12}\},\{F_{19}\},\\\{F_{20}\},\{F_{16}\},\{F_{17}\},\{F_{18}\}\}$. The first set $\{F_1,F_6,F_2,F_7\}$ clearly separates classes $1$ and $2$ from the other classes, where $F_1,F_6$ separates class $1$ from class $2$. The next subset $\{F_9,F_4,F_{14},F_{10},F_5,F_{15}\}$ segregates classes $3,4$ and $5$ amongst themselves. Since $F_{14},F_{15}$ were generated by randomly replacing $200$ negative values with positive values in $F_4,F_5$ respectively, they group up in together side-by side in the selected subset. Hence HOFS preserves the underlying structure of the generated features in presence of heterogeneity and provides meaningful feature ranking. Fig.~\ref{fig:hetero} shows the plot of Information gain according to the heterogeneous model and as predicted by HOFS.

\begin{table*}[h]
	\centering
	\begin{tabular}{|c|c|ccccc|ccccc|}
		\hline
		Sample ID & Class Label & $F_1$ & $F_2$ & $F_3$ & $F_4$ & $F_5$ & $F_6$ & $F_7$ & $F_8$ & $F_9$ & $F_{10}$\\\hline
		$1-100$    & $\textbf{1}$ & $\textbf{1}$ & $0$ & $0$ & $<0$ & $<0$ & $\textbf{1}$ & $0$ & $0$ & $<0$ & $<0$\\
		$101-200$  & $\textbf{1}$ & $\textbf{1}$ & $0$ & $0$ & $<0$ & $<0$ & $\textbf{1}$ & $\textbf{1}$ & $0$ & $<0$ & $<0$\\
		$201-300$  & $\textbf{2}$ & $\textbf{2}$ & $\textbf{1}$ & $0$ & $<0$ & $<0$ & $\textbf{1}$ & $\textbf{1}$ & $0$ & $<0$ & $<0$\\
		$301-400$  & $\textbf{2}$ & $\textbf{2}$ & $\textbf{1}$ & $0$ & $<0$ & $<0$ & $\textbf{1}$ & $\textbf{1}$ & $\textbf{1}$ & $<0$ & $<0$\\
		$401-500$  & $\textbf{3}$ & $0$ & $0$ & $1$ & $<0$ & $<0$ & $0$ & $\textbf{1}$ & $\textbf{1}$ & $<0$ & $<0$\\
		$501-600$  & $\textbf{3}$ & $0$ & $0$ & $1$ & $<0$ & $<0$ & $0$ & $0$ & $\textbf{1}$ & $>\textbf{0}$ & $<0$\\
		$601-700$  & $\textbf{4}$ & $0$ & $0$ & $0$ & $>\textbf{0}$ & $<0$ & $0$ & $0$ & $\textbf{1}$ & $>\textbf{0}$ & $<0$\\
		$701-800$  & $\textbf{4}$ & $0$ & $0$ & $0$ & $>\textbf{0}$ & $<0$ & $0$ & $0$ & $0$ & $>\textbf{0}$ & $>\textbf{0}$\\
		$801-900$  & $\textbf{5}$ & $0$ & $0$ & $0$ & $>\textbf{0}$ & $>\textbf{0}$ & $0$ & $0$ & $0$ & $>\textbf{0}$ & $>\textbf{0}$\\
		$901-1000$ & $\textbf{5}$ & $0$ & $0$ & $0$ & $>\textbf{0}$ & $>\textbf{0}$ & $0$ & $0$ & $0$ & $>\textbf{0}$ & $>\textbf{0}$\\\hline
	\end{tabular}
	\caption{Descriptions of significant features ($F_1-F_5$) and biased features ($F_6-F_{10}$) of
		synthetic data.}
	\label{table:synth}
\end{table*}

\section{Conclusion} \label{sec7}

Mutual Information (MI) defines a measurement of how informative the features are. Feature selection based on MI has been developed a lot over the past decade. However,the computation of global MI is a NP-Hard problem. Thus, most of those algorithms are forced to do a lower order estimation. We introduce an improved method (HOFS) to estimate the global MI by integrating incremental ICA algorithm. Our method clearly performs better than most owing to its ability to select feature subsets that jointly maximize MI, while keeping similar running times and computational complexity as the current approaches. We would like to extend our work to provide minimal cardinality feature subsets for a wide range of datasets including geometric and hyperspectral. 



\end{document}